\documentclass[10pt,twocolumn,letterpaper]{article}
\usepackage{iccv}
\usepackage{times}
\usepackage{epsfig}
\usepackage{graphicx}
\usepackage{amsmath}
\usepackage{amssymb}
\usepackage{float}
\usepackage{bbding}

\usepackage[breaklinks=true,bookmarks=false]{hyperref}

\iccvfinalcopy 

\def\iccvPaperID{2487} 

\ificcvfinal\fi

\begin{document}

\title{End-to-End Video Instance Segmentation via Spatial-Temporal Graph Neural Networks}

\author{Tao Wang\textsuperscript{1}, Ning Xu\textsuperscript{2}, Kean Chen\textsuperscript{1}, and Weiyao Lin\textsuperscript{1}\thanks{
Corresponding Author, Email: wylin@sjtu.edu.cn} \\
\textsuperscript{1}Shanghai Jiao Tong University, Shanghai, China\\
\textsuperscript{2}Adobe Research\thanks{
The paper is supported in part by the following grants: National Key Research and Development Program of China Grant (No.2018AAA0100400), National Natural Science Foundation of China (No. 61971277), and Adobe Gift Funding.}, San Jose, USA\\
{\tt\small wang\_tao1111@sjtu.edu.cn, nxu@adobe.com, ckadashuaige@sjtu.edu.cn, wylin@sjtu.edu.cn}
}

\maketitle
\ificcvfinal\thispagestyle{empty}\fi

\begin{abstract}
   Video instance segmentation is a challenging task that extends image instance segmentation to the video domain. Existing methods either rely only on single-frame information for  the detection and segmentation subproblems or handle tracking as a separate post-processing step, which limit their capability to fully leverage and share useful spatial-temporal information for all the subproblems. In this paper, we propose a novel graph-neural-network (GNN) based method to handle the aforementioned limitation. Specifically, graph nodes representing instance features are used for detection and segmentation while graph edges representing instance relations are used for tracking. Both inter and intra-frame information is effectively propagated and shared via graph updates and all the subproblems (\ie detection, segmentation and tracking) are jointly optimized in an unified framework. The performance of our method shows great improvement on the YoutubeVIS validation dataset compared to existing methods and achieves 36.5\% AP with a ResNet-50 backbone, operating at 22 FPS.
    
\end{abstract}

\section{Introduction}
\begin{figure}[t]
\begin{center}
   \includegraphics[width=0.97\linewidth]{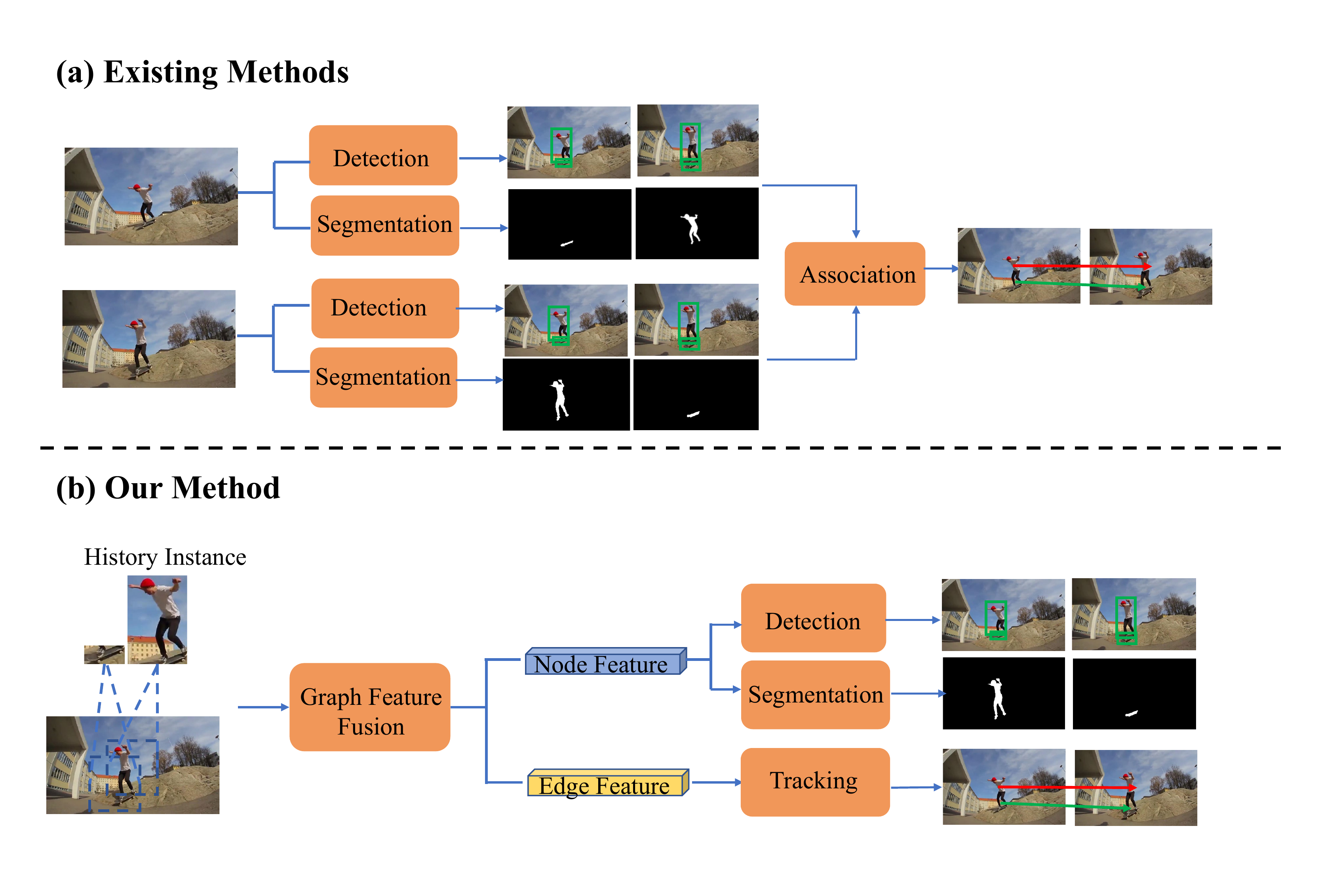}
\end{center}
   \caption{Comparison between existing methods and our methods. (a) Most existing methods solves VIS subproblems separately and ignore temporal feature fusion. (b) In our method, we use graph neural network to model the instance-level relations and fuse spatial-temporal features. Meanwhile, detection, segmentation and tracking results are jointly predicted based on graph features.}
\label{fig:brief}
\end{figure}

Video instance segmentation (VIS) is a challenging and fundamental task in computer vision. In the image domain, instance segmentation needs to simultaneously detect and segment object instances~\cite{hariharan2014simultaneous}, while in the time domain, the video instance segmentation \cite{yang2019video} is much more challenging since it also requires accurate tracking of objects across the entire videos.

tion problem for all the instances and candidates together, rather than focusing on single candidate one by one, avoiding some mismatch by utilizing information from related instances and candidates. 


Existing VIS methods typically adopt two different strategies to tackle the instance tracking task. The first strategy is to adopt the tracking-by-detection framework which first predicts candidate detection and segmentation frame by frame with a sophisticated image instance segmentation model, and then associates these candidates by classification or re-identification to generate mask sequences \cite{cao2020sipmask,luiten2019video,yang2019video}, as shown in Figure \ref{fig:brief}~(a). Its performance heavily relies on the qualities of image-level instance segmentation, the property of similarity metrics and the association strategy. The other strategy is to predict clip-level instance masks by propagate instance masks from central frame to the whole video clip and merge these clip-level sequences to generate video-level results \cite{bertasius2020classifying}. Such propagation process usually relies on some heuristics and thus a lot of post-processing and refinement operations are necessary.

In addition, both the strategies have one common limitation that their tracking is a separate step from detection and segmentation modules, forbidding useful information shared across different tasks. For example, a newly detected instance associated with an existing sequence whose class prediction label is ``dog" should be very helpful to predict the class of that instance, meaning that better tracking results should contribute to better detection or segmentation. Vice versa, improved detection and segmentation should also contribute to improved tracking. However, existing methods cannot easily leverage such benefits. 

Another common limitation of existing methods is that they ignore inter-frame and intra-frame instance relation information. The inter-frame instance relations refer to the relationship between instances of different frames while the intra-frame instance relations refer to the relationship between instances within the same frame. Such inter and intra-frame relations usually contain rich spatial-temporal information which is useful for all VIS tasks.  However, many existing methods directly detect and segment instance on single frame features.  They~\cite{bertasius2020classifying,cao2020sipmask,luiten2019video,yang2019video} also associate and track each new candidate with existing instance sequences independently, without having a global view from other candidate instances. A few recent methods~\cite{athar2020stem, bertasius2020classifying} have noticed the problem but they directly fuse the features from adjacent frames at the entire frame level, which could cause inaccurate information propagation and thus affect the accuracy negatively. In addition, they only leverage such information for detection and segmentation, but not for tracking.

We believe that both sharing information among different subtasks and extract inter and intra-frame information are crucial for the VIS task. Therefore, in this paper we propose a novel framework to achieve the two points at the same time, which is illustrated in Figure~\ref{fig:brief}~(b). Given a pair of reference frame and target frame, our method first builds a graph neural network (GNN) to connect the two frames where nodes represent instance candidates while edges represent instance relations. Then spatial-temporal features can be obtained via graph message passing. Our detection and segmentation branches which are inspired by recent anchor-free image detection and segmentation methods~\cite{zhou2019objects, tian2020conditional} leverage the updated node features to predict detection and segmentation of the target frame. While the tracking branch takes as input the updated edge features and performs binary classification to predict associations.  Through iterative GNN updates, the tracking information contained in graph edges and the detection/segmentation information contained in graph nodes are also shared. Our model is trained end-to-end and during inference, it is applied iteratively over pairs of consecutive frames to obtain detection, segmentation and tracking results simultaneously. 

Another novelty of our method is the segmentation branch. Given that the GNN features are high-level which may not contain enough shape information which are more useful for mask prediction, we propose a novel mask-information propagation module to warp the low-level instance shape features of the reference frame to the target frame. By using the warped features with the shape feature of the target frame to learn the mask head, our mask prediction head can achieve better segmentation results. 


We train and evaluate our method on YoutubeVIS dataset and our model with a ResNet-50 backbone achieves $36.5\%$ AP which is superior than most existing methods. Our method is also very efficient given its unified framework, which operates at a 22 FPS speed.

In summary, our method has four main contributions:
\begin{itemize}
\item We present a novel and efficient GNN-based framework for VIS to simultaneously detect, segment and associate instances.
\item We propose to leverage inter-frame and intra-frame features via feature aggregation in GNN, which is proved to be effective for all VIS subtasks.
\item We propose a mask-information propagation module to use the historical shape information for more accurate mask prediction.
\item We evaluate our method on the benchmark dataset and achieve competitive results compared to existing methods.
\end{itemize}

The rest of our paper is organized as follows. In Section \ref{section:related}, we briefly introduce related tasks and state the difference between our method and existing VIS methods. In Section \ref{section:approach} we describe our approach in detail. Experiments results are presented in Section \ref{section:experiments} and we conclude the paper in Section~\ref{section:conclusion}.


\begin{figure*}
\begin{center}
\includegraphics[scale=0.282]{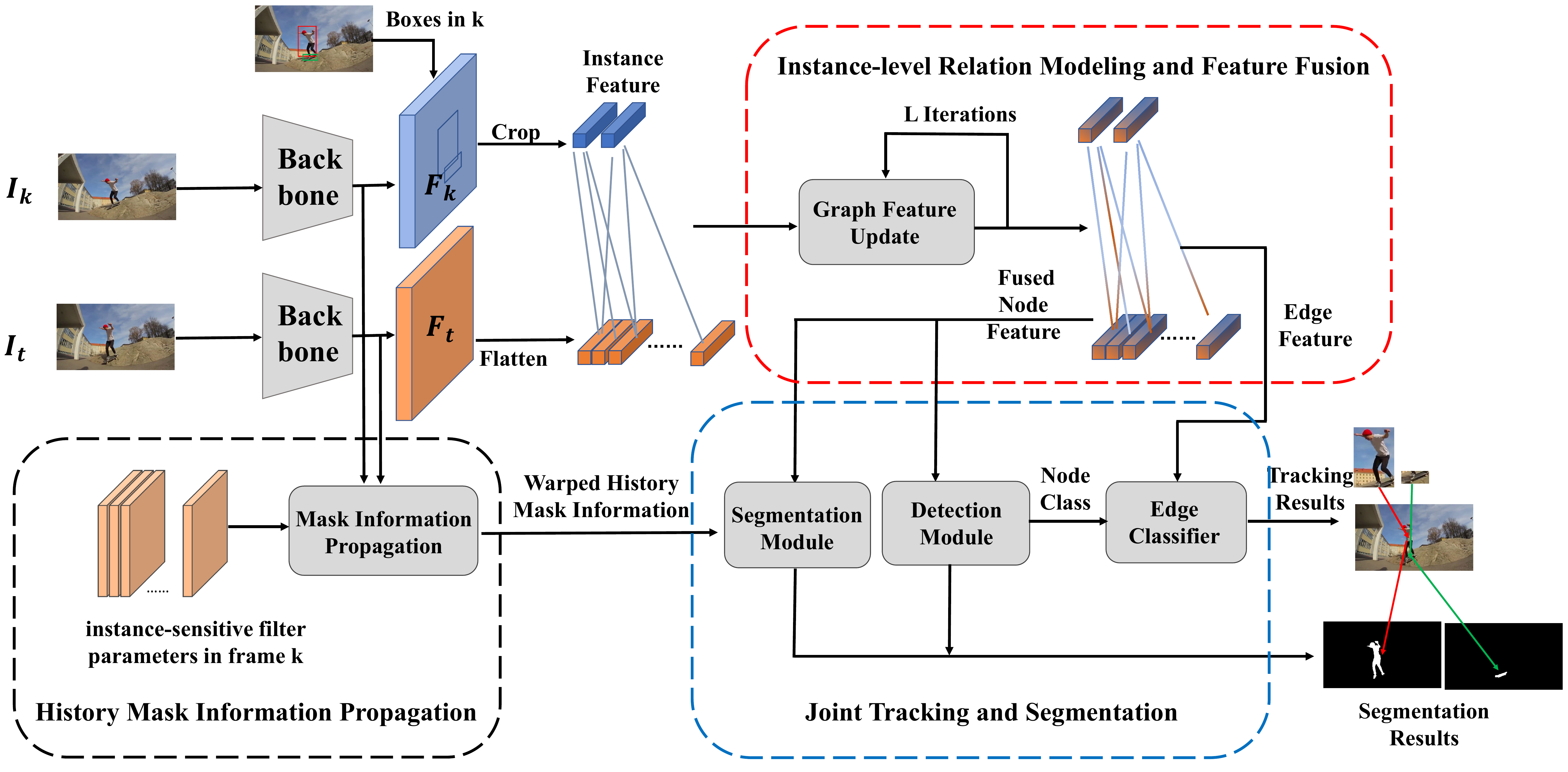}
\end{center}
   \caption{The overview of our framework. Given a reference frame $I_k$ and a target frame $I_t$, our method first construct a graph between the two frames and perform message passing to obtain aggregated spatial-temporal features for the target frame, which is shown in the red box. Then, our method predicts the detection and segmentation results of the target frame based on graph node features and predict tracking associations based on graph edge features, which is shown in the blue box. In addition, our segmentation branch has a mask-information propagation module (shown in the black box) to propagate the instance shape information of the reference frame to the target frame, which can improve the segmentation accuracy. All the modules are trained end-to-end and our method handles VIS detection, segmentation and tracking in an unified framework. For detailed structure of individual modules, please refer to Figure~\ref{fig:jointtracking} and~\ref{fig:onecol}.}   
   
\label{fig:framework}
\end{figure*}

\section{Related Work}\label{section:related}

\textbf{Image Instance Segmentation:} The aim of instance segmentation is to group pixels into different object instances to get classes, segmentation masks and unique identities of objects\cite{hariharan2014simultaneous}. Most of the instance segmentation can be divided into two-stage methods and one-stage methods. Two-stage methods always generate detection bounding boxes first and predict masks on these boxes. Mask R-CNN \cite{he2017mask} is still the mostly widely used two-stage framework in image instance segmentation and video instance segmentation, which generate object proposal by region proposal network and use two seperate branches to predict bounding boxes and segmentation masks on ROI features obtained from proposal areas on the feature map. Mask Scoring R-CNN \cite{huang2019mask} designs a MaskIOU branch to evaluate the accuracy of mask, rather than directly using the classification score in Mask R-CNN. PANet \cite{wang2019panet} use bottom-up path augmentation, adaptive feature pooling, and fully-connected fusion to improve the accuracy of segmentation.
In one-stage instance segmentation methods, masks are directly predicted from images without generating a large amount of proposals first. PolarMask \cite{xie2020polarmask} uses 16 rays to represent a mask and add these representation on FCOS \cite{tian2019fcos} framework to achieve instance segmentation. CondInst \cite{tian2020conditional} introduces conditional convolution into FCOS and uses dynamic instance-sensitive convolution filters to represent the masks. YOLACT \cite{bolya2019yolact} uses a group of templates and coefficients to represent masks and designs a fully convolutional network to predict them.  

\textbf{Video instance segmentation:} The video instance segmentation task is more challenging than video object segmentation\cite{li2020delving}, which requires classifying, segmenting instances in each frame and linking
the same instance across frames \cite{yang2019video}. It can be applied in many other computer vision fields, including human behavior analysis\cite{lin2020human} and object position estimation\cite{zhang2021regional}. The baseline method MaskTrack R-CNN \cite{yang2019video} adds a embedding branch on Mask R-CNN \cite{he2017mask} and uses a historical instance feature queue to associate instance masks across frame. Maskprop \cite{bertasius2020classifying} propagates masks in previous frames to predict their positions and shapes in the later frames. VisTR \cite{wang2020end} introduces transformers and uses queries on encoding features to obtain instance sequence from video clips. SipMask \cite{cao2020sipmask} designs a novel image instance segmentation method and uses the same association strategy of MaskTrack R-CNN \cite{yang2019video} to link instance masks across frames. STEm-Seg \cite{athar2020stem} models video clips as 3D space-time volume and generate instance sequences by clustering pixels based on space-time embedding. {We found that most of these methods tracks and segments each instance separately, without extracting and utilizing relation information between instances, while our method use instance-level relations extracted by graph neural network to guide the feature fusion and learn joint detection, segmentation as well as tracking in one framework, which is not only more efficient but also can improve the accuracy of all VIS subproblems.}

\textbf{Graph Neural Networks for Relation and Tracking:} Graph neural network was first proposed to process data with a graph
structure using neural networks \cite{gori2005new}. In GNN, 
a graph with nodes and edges relating each other is constructed and features on nodes and edges will be updated based on relations. GNN has already been used in object detection and multi object tracking to improve the accuracy. In object detection, GNN is often used to fuse the features of related objects or proposals to increase the detection accuracy \cite{wang2020joint,chen2020relation, shi2020point, xu2020cross}. As for
MOT, most of the methods formulates data association as an edge classification problem using GNNs \cite{braso2020learning, jiang2019graph, li2020graph, weng2020gnn3dmot}, where
each node represents a detection or a sequence of an object and each edge denotes the similarity between detection and sequences.
After aggregating node features and edge features based on object relations, the final association results can be obtained by classifying the edges. However, in MOT, tracking is usually a separate step from other tasks such as detection. While in our method graph is leveraged to fuse spatial-temporal features which are useful for not only tracking but also other tasks such as detection and segmentation .



\section{Approach}\label{section:approach}
To simultaneously detect, segment and associate instances in an unified framework and leverage useful inter-frame and intra-frame features, we introduce GNN to model the instance-level relations. Specifically, the feature maps of the current frame are aggregated with instance features of the previous frame via the GNN (Sec.~\ref{subsection:relation}) \cite{wang2020joint}. After some iterations of graph convolution and updates, the aggregated node features of the current frame will be used to predict detection (Sec.~\ref{subsection:detection}) and segmentation results (Sec.~\ref{subsection:propagation}), while edge features will be used to predict tracking results (Sec.~\ref{subsection:tracking}). In addition, to better leverage the mask predictions of previous frame, we introduce another novel module (Sec.~\ref{subsection:propagation}) to warp instance-sensitive mask filters of previous frame to the current frame to obtain more accurate mask prediction. The whole framework of our method is shown in Figure~\ref{fig:framework}. 

\subsection{Problem Definition}\label{subsection:definition}
 During training, given a pair of frames $I_k$ and $I_{t}$ of a video at time $k$,~$t$, and existing instance sequences at $k$ including bounding box detections denoted as $B_{k}=\{B_{k}^1,B_{k}^2,...,B_{k}^m\}$ and masks denoted as $M_{k}=\{M_{k}^1,M_{k}^2,...,M_{k}^m\}$ where $m$ is the number of existing instances, we aim at obtaining a set of instance candidates at frame $t$ with detections $B_{t}=\{B_{t}^1,B_{t}^2,...,B_{t}^n\}$ and masks $M_{t}=\{M_t^1,M_t^2,...,M_t^n\}$ where $n$ is the number of new instances which can be different from $m$. In addition, we need to assign the $n$ instances to the existing $m$ instance sequences or initialize unmatched candidates to new instance sequences. During inference, we iterate the above process over pairs of consecutive frames (\ie $I_{t-1}$ and $I_{t}$) to obtain the prediction of a whole video sequence. Our method is also able to handle the case where an existing instance sequence is occluded at frame $I_{t-1}$ but reappears at frame $I_{t}$, which will be detailed in Sec.~\ref{subsection:test}.


\begin{figure}[t]
\begin{center}
   \includegraphics[width=0.97\linewidth]{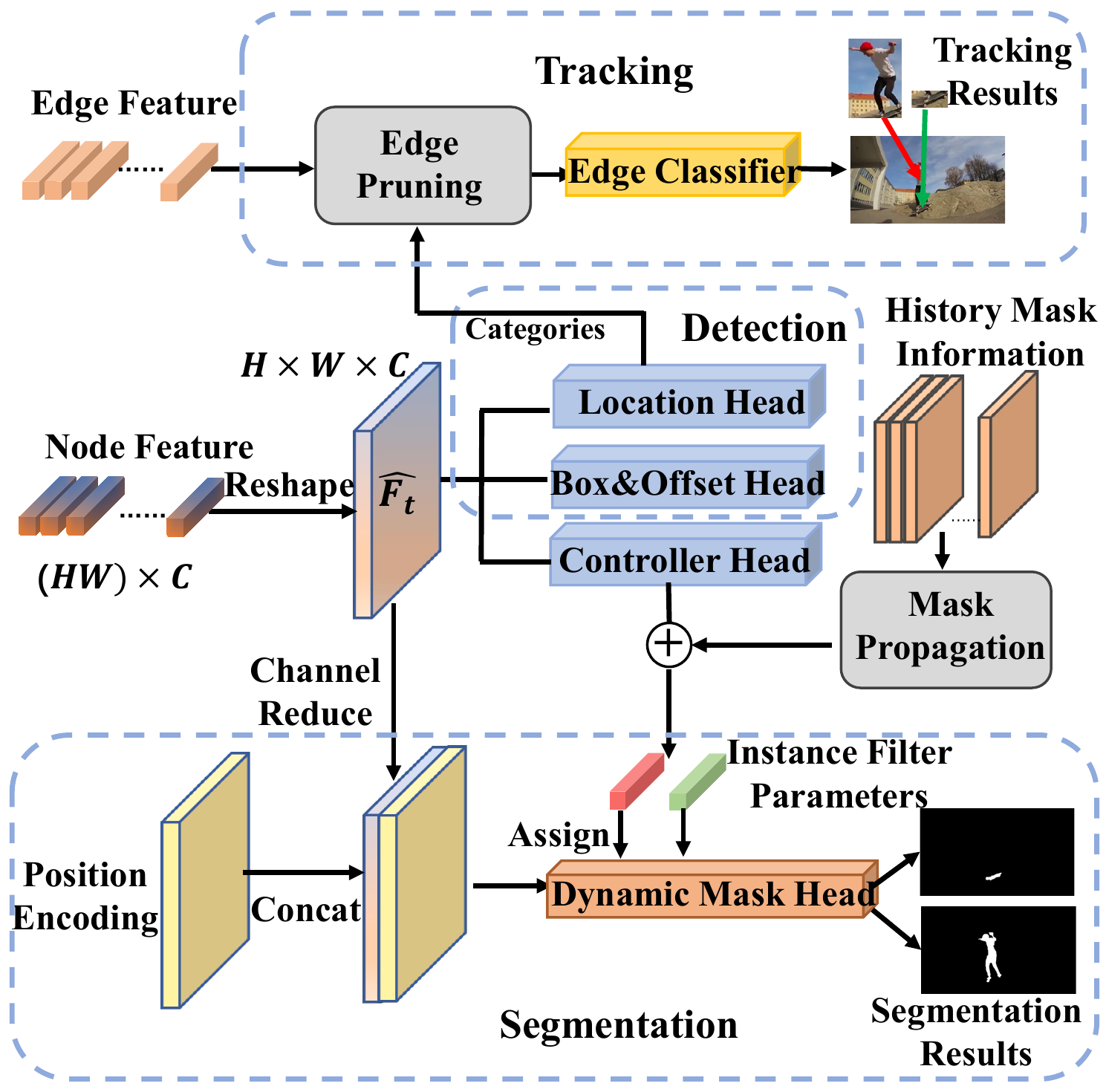}
\end{center}
   \caption{The detailed structure of our detection, segmentation and tracking branches. }
\label{fig:jointtracking}
\end{figure}     
\subsection{Spatial-Temporal Feature Fusion via GNN}\label{subsection:relation}
Different from images, video contains rich information across different frames and different spatial locations. Given the feature maps $F_k,F_t\in\mathcal{R}^{H\times W\times D}$ of frame $k$ and $t$ extracted by a CNN encoder (\eg ResNet-50), our method aims to produce an aggregated feature map $\hat{F}_{t}$ for frame $t$ which fuses both inter-frame ($F_k$) and intra-frame ($F_t$) features for better prediction of the detection and segmentation subproblems than using the single-frame feature $F_t$. In addition, we also want to predict tracking results simultaneously. All the goals are achieved by a GNN which is explained as follows.

\textbf{Graph Construction:} Our graph $G(V,E)$ consists of a node set $V$ and an edge set $E$. The node definition is different for frame $k$ and $t$ since at frame $k$ we already know the instance candidates while at frame $t$ we do not. Therefore, for frame $k$ we define each instance candidate as a node. The node feature is defined as the cropped feature of $F_k$ by ROIAlign \cite{he2017mask} and further downsampled and flattened to one $D$-dimensional feature vector (as shown in Fig.~\ref{fig:framework}). For frame $t$, inspired by anchor-free image detection methods such as CenterNet~\cite{zhou2019objects}, we define each pixel location $(x,y)$ of $F_t$ as a node and $F_t(x,y)\in \mathcal{R}^{D}$ as its node feature. Despite the different definition, we use $h^v_i$ to denote the feature of node $i$ in set $V$.



An edge $e_{ij}$ in set $E$ represents an undirected connection between node $i$ and $j$ and $h^e_{ij}$ denotes the edge feature which is initialized as $|h^v_i-h^v_j|$. Our graph is sparsely connected with the following two criteria. First, only nodes across frames can be connected (as shown in Fig.~\ref{fig:framework}). Second, for an instance node at frame $k$ whose center is $(x,y)$, it is only connected to the nodes at frame $t$ whose distances to $(x,y)$ is smaller than a threshold (\eg a $w\times{w}$ window around the center).

There are three motivations for our choice. The first reason is the memory and computation concern since a dense graph will cost too many resources \cite{wang2020joint}. Secondly, our method uses edge features to predict tracking results and thus only cross-frame edges are meaningful. Thirdly, the displacement of an instance between nearby frames is usually local. It also should be noted that although our graph only has inter-frame connections, the iterative graph updates can still aggregate useful intra-frame features into the fused feature map $\hat{F}_{t}$, which will be explained next.


\textbf{Graph Feature Update:}
We follow the mechanism of previous message passing networks \cite{gilmer2017neural, braso2020learning} to update our graph. Specifically, at each iteration, the node and edge features are propagated and updated in the following order: 1) edge features are updated by its two endpoints, 2) node features are updated by its connected edges, the process of which can be represented as follows.
\begin{equation}
h_{ij}^{e(l)}=N_e([h_{ij}^{e(l-1)},h_{i}^{v(l-1)},h_{j}^{v(l-1)}])
\end{equation}
\begin{equation}
h_{i}^{v(l)}=h_{i}^{v(l)}+\sum_{j | e_{ij} \in E}N_v([h_{ij}^{e(l)},h_{i}^{v(l-1)}])
\end{equation}
where $N_e$ and $N_v$ denote two learnable functions which are two multilayer perceptrons (MLP) in our experiments. $[\cdot]$ denotes the concatenation operation. After total $L$ iterations, the updated node feature at frame $t$ are reshaped to $\mathbb{R}^{H\times{W}\times{D}}$ to create the aggregated feature $\hat{F}_{t}$ which will be used to predict detection and segmentation results while the updated edge features will be used to predict tracking results.

It is worth noting that as long as the graph update iterations $L\ge 2$, the updated node and edge features will contain both inter-frame and intra-frame features even though our graph has no intra-frame connections. Since a node will receive the information of its nearby nodes of the same frame after their connected common node of the other frame aggregates all their information after the first iteration. Such design enables our graph to be sparsely connected and thus requires less computation resources while still manage to extract useful spatial-temporal features.

\subsection{Detection With Node Features}\label{subsection:detection}
We use the aggregated spatial-temporal feature map $\hat{F}_{t}$ for the detection branch. We follow the anchor-free image detection method CenterNet \cite{zhou2019objects} to predict object categories and locations by finding their center points and bounding box sizes. Specifically, $\hat{F}_{t}$ is fed into a three-convolutional head to obtain the location heat map, size map and coordinate-refinement map. The location heat map estimates instance center points and their categories. The size map estimates bounding box sizes. The refinement map helps refine the locations of center points. The loss of our detection branch follows that in CenterNet~\cite{zhou2019objects} which consists of the center-point loss, size loss and offset loss:
\begin{equation}
L_{\text{det}}= L_\text{center}+\lambda_{\text{size}}L_{\text{size}}+\lambda_{\text{offset}}L_{\text{offset}}
\end{equation}
where $\lambda_{\text{size}}$ and $\lambda_{\text{offset}}$ balances the three losses. 
\subsection{Segmentation Branch}\label{subsection:propagation}
\begin{figure}[t]
\begin{center}
   \includegraphics[width=0.97\linewidth]{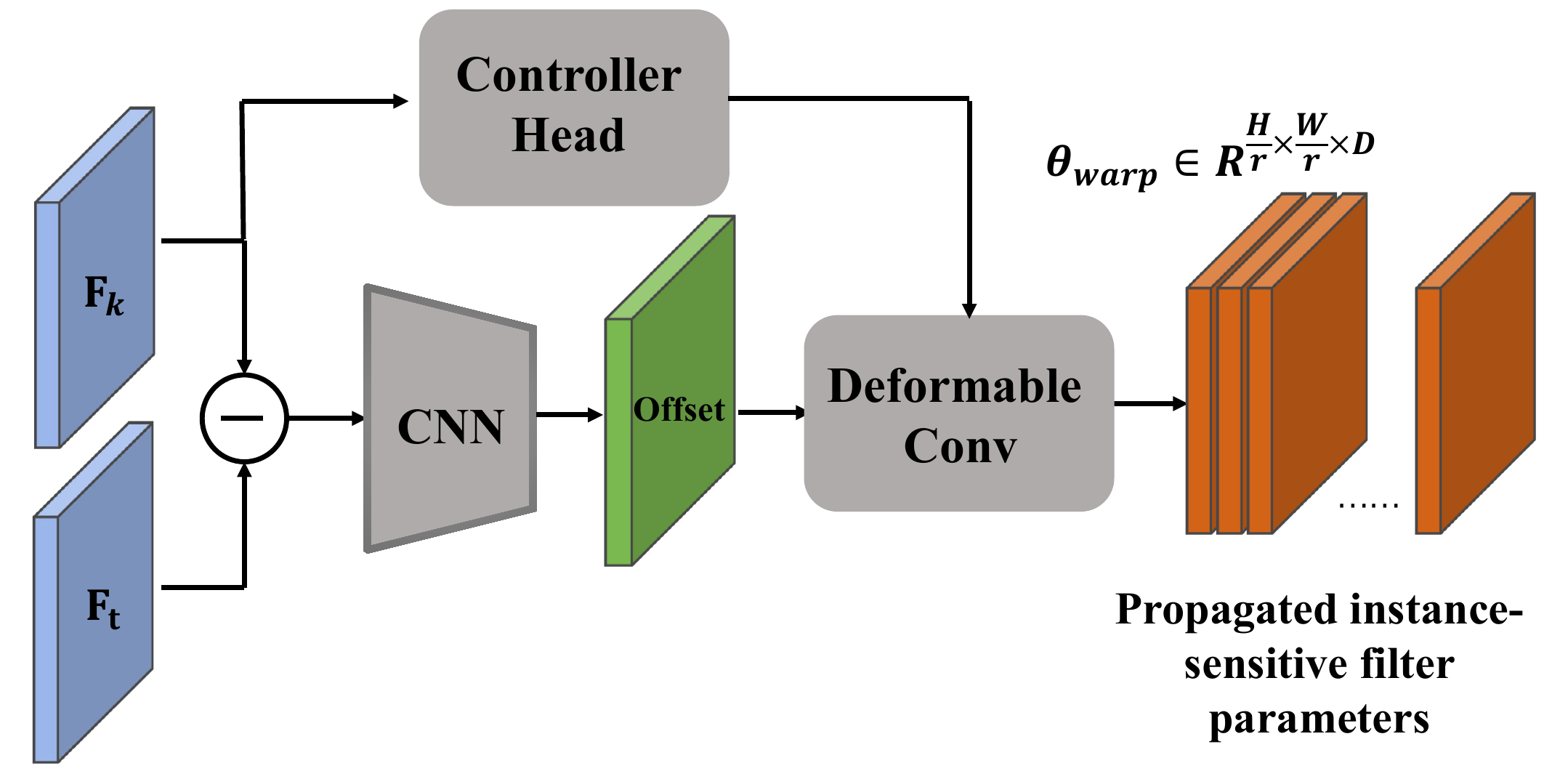}
\end{center}
   \caption{The structure of our warping module within the segmentation branch. The module uses deformable convolution \cite{dai2017deformable} to warp historical instance convolution kernels and use them together with those predicted by the controller head to predict instance masks. }
\label{fig:propagation}
\label{fig:onecol}
\end{figure}
Inspired by recent image instance segmentation method~\cite{tian2020conditional} which predicts dynamic convolution filters for each instance candidate, our segmentation branch also has a controller head to predict such instance-sensitive convolution filters  $\theta_t\in\mathcal{R}^{{H}\times{W}\times{D}}$. The convolution filter $\theta_t(x,y)\in\mathcal{R}^D$ at an instance location $(x,y)$ can then be used as the $1\times1$ convolution kernels in the mask head (\ie a fully-convolutional network (FCN)) to predict the corresponding instance mask (as shown in Figure~\ref{fig:jointtracking}). One difference is that our controller head leverages the spatial-temporal features $\hat{F}_t$ which contains richer information than that of a single frame.

Furthermore, in our setting, the instance-sensitive kernels $\theta_k$ at the reference frame $k$ which represent historical instance mask and shape information are also available, which could be useful to refine the segmentation results at frame $t$. Therefore we propose another novel module to warp $\theta_k$ to the current frame based on the feature differences of the two frames. 
Specifically, as shown in Figure \ref{fig:propagation}, given the two feature maps $F_k$ and $F_t$, their difference is fed into a CNN to estimate a group of offsets. The offsets are reshaped to two $3\times3$ filters for x and y axis and then used as deformable convolution parameters. The input of the deformable convolution is $\theta_k$ while the output is the warped kernels $\theta_{warp}$ for the current frame. We believe that $\theta_{warp}$ can also be used to decode the instance masks in the current frame. Thus, we also use it to guide the learning of the mask branch. Our warping module shares a similar spirit to the alignment operation in MaskProp~\cite{bertasius2020classifying} but has two major differences. First, our warping operates at the feature level while MaskProp warps at the mask level. Second, our method simultaneously warps all instance filters while MaskProp warps one instance mask at a time, and thus is less efficient than ours.

Finally, to predict an instance mask at location $(x,y)$, our mask head takes as inputs the instance-sensitive kernels $\theta_t(x,y)$ and $\theta_{warp}(x,y)$, the spatial-temporal feature map $\hat{F}_t$ as well as the position encoding map $P_{x,y}$~\cite{tian2020conditional} as follows.
\begin{equation}
M_{x,y}=\text{MaskHead}([\hat{F}^d_t,P_{x,y}];\theta_t(x,y))
\end{equation}
\begin{equation}
M_{x,y}^{'}=\text{MaskHead}([\hat{F}^d_t,P_{x,y}];\theta_{warp}(x,y))
\end{equation}
where $[\cdot]$ denotes the concatenate operation, $\hat{F}^d_t$ is generated from  $\hat{F}_t$ by $1\times{1}$ convolution to reduce the channel dimension to 8. $P_{x,y}$ is a two-channel position map which encodes the relative x-axis and y-axis offsets from $(x,y)$ to the other locations. Our MaskHead operation is the same as~\cite{tian2020conditional}. In addition, we define the loss function for the segmentation branch $L_{\text{mask}}$  as the dice loss \cite{milletari2016v} between predicted instance masks and ground truth masks. It can be seen that our segmentation branch leverages two types of spatial-temporal information, \ie both the aggregated semantic feature via GNN and the historical shape-related kernels from previous frame.

\subsection{Tracking With Edge Features}\label{subsection:tracking}
Our tracking branch utilizes the graph edges to associate existing tracks or initialize new tracks. However, since many graph nodes at frame $t$ are classified as background in the detection branch, it is redundant to still keep these edges and thus they are removed prior to the tracking prediction. Moreover, some graph nodes at frame $t$ have different class predictions with their connected graph nodes at frame $k$. These are also meaningless edges since no instance can change its class. Therefore we also remove this type of edges from the graph. The remaining edges are kept for tracking prediction. 
Specifically, for each remaining edge $e_{ij}$, its final edge feature $h^{e(L)}_{ij}$ is fed into a binary classifier to predict the association relationship between node $i$ at frame $k$ and node $j$ at frame $t$. A positive prediction indicates the two nodes belong to the same instance and vice versa. Therefore the loss for our tracking branch is defined as:
\begin{equation}
L_{\text{edge}}\!=\!\frac{-1}{|E_{\text{prune}}|}\!\sum_{e_{ij} \in 
E_{\text{prune}}}\!y_{ij}\log(\hat{y}_{ij})\!+\!(1\!-\!y_{ij})\log(1-\hat{y}_{ij})
\end{equation}
where $\hat{y}_{ij}$ is the prediction for edge $e_{ij}$ and $E_{\text{prune}}$ denotes the pruned edge set after removing the redundant edges.

A positive prediction indicates that the node $j$ is the extension of node $i$ in current frame and thus should be associated to the same track. For those instances with multiple matched edges, our method only keeps the edges with the highest classification score and remove the remaining ones. While if an instance of frame $t$ is not matched to any existing sequence after edge classification, a new track based on this instance will be initialized.

\subsection{Training and Testing}\label{subsection:test}
\textbf{Training:} We randomly select two frames $I_k$ and $I_t$ from a video sequence as the reference and target frame, where the time gap between $k$ and $t$ should be smaller than 5. We use the ground truth detections of the reference frame to extract its node features. We use the controller head to obtain the instance-sensitive convolution filters of the reference frame. 
The total loss function of our model can be written as: 
\begin{equation}
L_{\text{total}}=\lambda_1L_{\text{det}}+\lambda_2L_{\text{mask}}+\lambda_3L_{\text{edge}}
\end{equation}
The loss function is applied on the target frame while the segmentation loss $L_{\text{mask}}$ is also applied on the reference frame.

\textbf{Testing:} Given a video sequence, our method follows the online strategy to processes a pair of consecutive frames  $I_{t-1}$ (reference frame) and $I_t$ (target frame) iteratively until the end of the video sequence. For the first frame, since there is no reference frame, we only use the plain detection and segmentation branches (\ie no spatial-temporal features nor kernel warping module) to get its instance predictions. The full tracking results can be easily obtained by connecting the tracking prediction of each pair of consecutive frames. However, it is possible that an instance in the reference frame $I_{t-1}$ is not matched to any instances in the current frame $I_t$, indicating that the instance is occluded in $I_t$. In this case, our method puts the unmatched instance identity together with its cropped node feature $h_i^v$ as well as its instance-sensitive convolution filters $\theta_{t-1}(x,y)$ in a memory. When proceeding to the next pair of frames $I_t$ and $I_{t+1}$, we will add the unmatched node feature as an existing instance node at frame $t$. To make the assumption hold, we only keep the unmatched instance in the memory within the time interval $\Delta t$, which is set to 7 in our experiments. 


\section{Experiments}\label{section:experiments}
\subsection{Datasets and Metrics}
In this section, we conduct our experiments on the YouTubeVIS \cite{yang2019video} dataset, which contains 2238 training, 302 validation and 343 test video clips. The dataset is
annotated with 40 categories, object bounding boxes, segmentation masks and instance identity labels.

Video instance segmentation is evaluated by the metrics of AP (Average Precision) and AR (Average Recall). To achieve a good performance, the method must accurately segment instances and associate instances across frames correctly at the same time.

\subsection{Implementation Details}
\textbf{Network:} We use ResNet-50 \cite{he2016deep} as our feature extraction backbone and build our detection branches as same as CenterNet \cite{zhou2019objects}. We also use FPN module in our framework. Ground truth instances are assigned to different FPN levels according to their sizes. Relation modeling and feature fusion are processed on each output feature map from FPN. Following the settings of segmentation module in CondInst \cite{tian2020conditional}, we use three convolution layers as our dynamic mask FCN head, thus, the dynamic controller head have to output 169-channel parameter maps including ${H}\times{W}$ instance-sensitive filter parameter vectors for each potential instance. The kernel size of deformable convolution \cite{dai2017deformable} used in propagation branch is $3\times{3}$.

\textbf{Training:} We use backbone model pretrained on COCO dataset and train the whole network on YouTubeVIS \cite{yang2019video} dataset on 2 Titan RTX GPUs for 24 epochs. The initial learning rate is 1e-2 and batch size is 16. The learning rate is reduced by a factor of 10 at epoch 16 and 22. We also use multi-scale data augmentation during training.

\subsection{Ablation Experiments}
In this section we conduct some ablation experiments to study the improvement from each module and find the best hyper-parameters for them.

\textbf{Iterations of GNN Update:} To study the impact from the iterations of the GNN feature update, we set the update iterations from 1 to 4 and evaluate the model on YoutubeVIS \cite{yang2019video} validation dataset. From Table \ref{tab:GNN}, we can observe that AP on validation dataset increases when update iterations increase from 1 to 3, because each pixel in feature map can obtain more relevant information from its neighboring instances from last frame. However, when iterations is larger than 3, AP drops. {The reason is that when layer number of graph neural network increases, the node and edge features will converge to the average feature, making the network lose its expressive power \cite{oono2019graph}. So the classification and segmentation heads are not able to predict accurate results from these features that contain less information.}

\begin{table}
\begin{center}
\resizebox{0.45\textwidth}{!}{
\begin{tabular}{|c|ccccc|}
\hline
Iterations & $AP$ & $AP_{50}$ & $AP_{75}$ & $AR_1$ & $AR_{10}$\\
\hline
1  & 32.1 & 53.9 & 34.1 & 31.7 & 36.5 \\
2  & 35.3 & 56.4 & 37.8 & 33.9 & 38.9 \\
3  & 36.5 & 58.6 & 39.0 & 35.5 & 40.8 \\
4  & 36.0 & 58.1 & 38.8 & 35.0 & 40.3 \\
\hline
\end{tabular}
}
\end{center}
\caption{Performance on YoutubeVIS dataset when the number of update iterations in GNN increases from 1 to 4.}\label{tab:GNN}
\end{table}

\begin{table}
\begin{center}
\resizebox{0.47\textwidth}{!}{
\begin{tabular}{|cccc|}
\hline
Feature Fusion & Association Strategy & Propagation& $AP$ \\
\hline
No fusion & Embedding & \XSolidBrush & 31.6 \\
Concat & Embedding & \XSolidBrush & 32.1 \\
Relation & Offset & \XSolidBrush & 32.6 \\
Relation & Embedding & \XSolidBrush & 33.4 \\
Relation & Edge Feature & \XSolidBrush & 34.8 \\
\hline
No fusion & Embedding & \checkmark & 32.0 \\
Relation & Embedding & \checkmark & 33.9 \\
Relation & Edge Feature & \checkmark & 35.2 \\
\hline
\end{tabular}
}
\end{center}
\caption{Ablation experiment results on feature fusion, tracking strategy and mask information propagation module. Multi-scale training is not used. ``Relation'' and ``Edge features'' denote our feature fusion and tracking strategy based on gnn, while ``Embedding" and ``Offset" denote tracking strategy following other VIS or MOT method.}\label{tab:ablation}
\end{table}

\textbf{Feature Fusion Strategy:} We argue that feature fusion based on relation modeling is important for improving the accuracy of video instance segmentation.
To prove it, we use different feature fusion strategies in our ablation experiments. From Table \ref{tab:ablation}, we find that models using fusion strategy based on relation modeling get much higher AP than models without fusion. The reason is that the former outputs more accurate masks and tracking relations based on received temporal information. The fusion strategy based on relation modeling also outperforms the strategy that fuses feature maps of two frames directly by using the concatenation of them as input. This is because, compared to the scene-level information, the instance-level relations and instance corresponding information are more helpful while they are ignored by methods that directly use CNN to fuse features.

\textbf{Joint Detection, Segmentation and Tracking:} To prove that jointly learning detection, segmentation and tracking based on instance-level relations is beneficial for improving the performance on video instance segmentation task, we conduct the experiments using different association strategies. {For ``Embedding" association strategy in Table \ref{tab:ablation}, we follow the strategy in MaskTrack R-CNN \cite{yang2019video} by adding an extra head to generate an embedding feature and using this embedding to associate the most similar candidates with corresponding instances. The embedding head is trained with classification loss as same as that in MaskTrack R-CNN. As for ``Offset" association strategy, following CenterTrack \cite{zhou2020tracking}, we predict center point offsets across frames for each instance and use these offsets and IoU to match instances with candidates.} We compare the performance of these two association strategies with our edge features based association strategy. We find that association strategy based on edge features outperforms both of them. The association based on offsets performs worse because YoutubeVIS \cite{yang2019video} dataset is annotated 1 frame per 5 frame from videos, while offset tracking relies on high fps videos. Meanwhile, different from MOT situation where the categories only include person and vehicle, which means the motion pattern of objects is relatively easy to learn, in YoutubeVIS, there are various classes of objects and it is hard to model their motion patterns. As for association based on embedding, though the embedding head is trained together with segmentation head, it treats the association between two frames as a N-classes classification problem, where N is the instance in the previous frame. However, N varies in each frame and it is hard to learn the classifier. Moreover, each of the candidate in current frame is classified separately, without using information from other candidates. Relation modeling based on GNN utilizes the features from all instances and candidates and a binary classifier for edges is also more robust, so it performs much better than embedding-based association.

\textbf{Mask Information Propagation Branch:} Most of the VIS methods utilize temporal information by feeding fused feature maps into prediction heads, but ignores the concrete mask-level information provided by historical instance masks. From Table \ref{tab:ablation}, we find that adding a mask information  propagation branch is helpful to improve AP value. The reason is that by merging the warped filter information from last frame, the mask head not only segments instances based on fused feature map $\hat{F}_t$, but also utilizes the instance-level mask information, motion and alignment information from the previous frame, which makes it learn the shape of instances more robustly. In addition to fusing temporal information via GNN, using the propagation branch will offer more detailed shape information for the mask prediction head.

%
\begin{figure}[t]
\begin{center}
\includegraphics[scale=0.265]{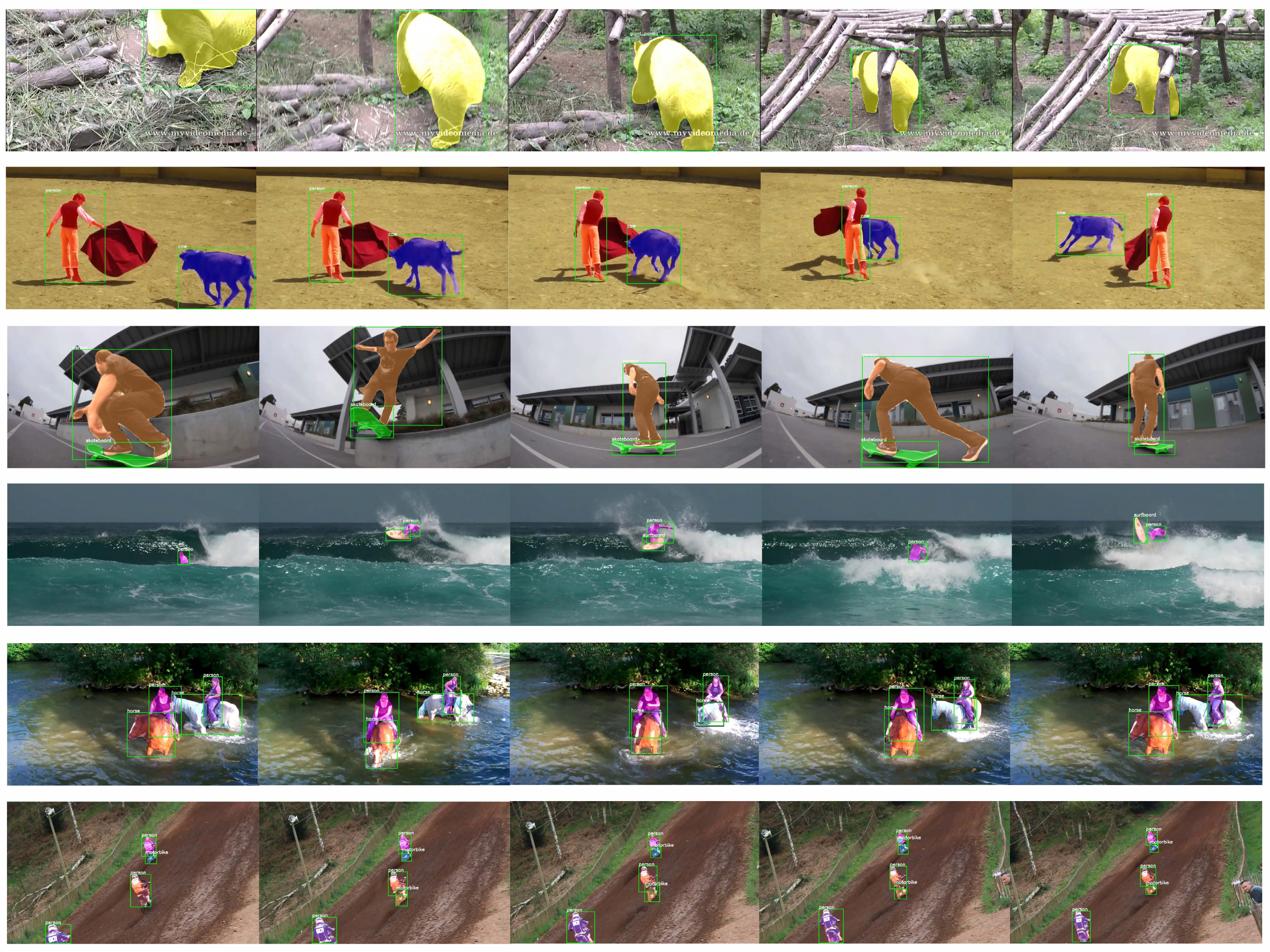}
\end{center}
   \caption{Visualization results of our method on YoutubeVIS dataset. Best viewed in color.}
\label{fig:results}
\end{figure}

\subsection{Main Results}
We compare our method with existing VIS methods in Table \ref{tab:compare}. We use metrics $AP$, $AP_{75}$ and $AP_{50}$ to evaluate these methods. To eliminate the effect from backbones, all the methods in table use ResNet-50 \cite{he2016deep} as the backbone. We can observe that our method achieves higher AP value than most of the existing VIS methods. Especially, the AP value of our method is 6.2\% higher than baseline method MaskTrack R-CNN \cite{yang2019video} and 2.1\% higher than the recent method VisTR which is based on transformer. 
Note that AP of our method is lower than that of MaskProp \cite{bertasius2020classifying}. We believe it is caused by the complicated post-processing steps and a stronger network, while our method deals with the segmentation, detection and tracking in a single framework, and thus has much faster FPS than~\cite{bertasius2020classifying} (22~\vs~2).
Some visualization results of our method on the YouTube-VIS
validation dataset are shown in Figure \ref{fig:results}.
Images in each row are sampled from the same video and the instance masks with the same color belong to the same object. We observe that our method can track and segment well in various situations, even some instances are occluded or overlapped. 
\begin{table}
\begin{center}
\resizebox{0.47\textwidth}{!}{
\begin{tabular}{|l|ccccc|}
\hline
Method & $AP$ & $AP_{50}$ & $AP_{75}$ & $AR_1$ & $AR_{10}$\\
\hline
DeepSORT\cite{wojke2017simple} & 26.1 & 42.9 & 26.1 & 27.8 & 31.3 \\
FEELVOS\cite{voigtlaender2019feelvos} & 26.9 & 42.0 & 29.7 & 29.9 & 33.4\\
OSMN\cite{yang2018efficient} & 27.5 & 45.1 & 29.1 & 28.6 & 33.1\\
MaskTrack R-CNN\cite{yang2019video} & 30.3 & 51.1 & 32.6 & 31.0 & 35.5 \\
MaskProp\cite{bertasius2020classifying} & 40.0 & - & 42.9 & - & - \\
STEm-Seg\cite{athar2020stem} & 30.6 & 50.7 & 33.5 & 31.6 & 37.1 \\ 
VisTR\cite{wang2020end} & 34.4 & 55.7 & 36.5 & 33.5 & 38.9\\
SipMask\cite{cao2020sipmask} & 33.7 & 54.1 & 35.8 & - & -\\
\hline
\textbf{VisSTG(ours)} w/o ms & 35.2 & 55.7 & 38.0 & 33.6 & 38.5 \\
\textbf{VisSTG(ours)} & 36.5 & 58.6 & 39.0 & 35.5 & 40.8 \\
\hline
\end{tabular}
}
\end{center}
\caption{The results of video instance segmentation on the YouTube-VIS validation dataset. We compare our method to other existing methods by metrics of $AP$, $AP_{50}$ and $AP_{75}$. All of the methods use ResNet-50 as backbone. Our method outperforms most of the existing methods. }\label{tab:compare}
\end{table}
\section{Conclusion}\label{section:conclusion}
We proposed an end-to-end video instance segmentation approach that simultaneously learns classification, detection, segmentation and tracking. To better utilize spatial-temporal information from previous frames and related instances, we use GNN to model the instance-level relations and guide the spatial-temporal information fusion. Detection and segmentation results are predicted from fused node features, while tracking results can be obtained from edge features via a binary edge classifier simultaneously. To further improve the segmentation accuracy, we add a propagation branch to obtain history mask-level information and use the warped instance-sensitive filters to predict masks by a dynamic mask head for each instance. Our method utilizes instance-level spatial-temporal information for all VIS subproblems, achieving good performance on YoutubeVIS validation dataset.

{\small
\bibliographystyle{ieee_fullname}
\bibliography{egbib}
}

\end{document}